\documentclass[letterpaper, 10 pt, conference]{ieeeconf}
\IEEEoverridecommandlockouts

\usepackage{cite}
\usepackage{amsmath,amssymb,amsfonts}
\usepackage{algorithmic}
\usepackage{graphicx}
\usepackage{textcomp}
\usepackage{xcolor}
\usepackage{tikz}
\usepackage{lipsum}
    
\usepackage[utf8]{inputenc}
\usepackage[english]{babel}

\usepackage[hyphens]{url}
\usepackage{hyperref}

\usepackage{amsthm}
\usepackage{booktabs}
\usepackage{multicol}
\usepackage{multirow}
\usepackage{makecell}
\usepackage{breakurl}
\usepackage{xspace}

\theoremstyle{definition}
\newtheorem{definition}{Definition}

\newcommand\copyrighttext{%
  \footnotesize \textcopyright 2021 IEEE.  Personal use of this material is permitted.  Permission from IEEE must be obtained for all other uses, in any current or future media, including reprinting/republishing this material for advertising or promotional purposes, creating new collective works, for resale or redistribution to servers or lists, or reuse of any copyrighted component of this work in other works.}
\newcommand\copyrightnotice{%
\begin{tikzpicture}[remember picture,overlay]
\node[anchor=south,yshift=10pt] at (current page.south) {\fbox{\parbox{\dimexpr\textwidth-\fboxsep-\fboxrule\relax}{\copyrighttext}}};
\end{tikzpicture}%
}

\begin{document}

\newcommand{\approach}[0]{DFDS\xspace}

\title{\LARGE \bf
Deep Information Fusion for Electric Vehicle Charging Station Occupancy Forecasting
}

\author{Ashutosh Sao$^{1}$, Nicolas Tempelmeier$^{2}$ and Elena Demidova$^{3}$% <-this % stops a space
\thanks{$^{1}$Ashutosh Sao is with L3S Research Center, Leibniz University Hannover, Appelstraße 9a, 30167 Hannover, Germany
        {\tt\small sao@L3S.de}}%
\thanks{$^{2}$Nicolas Tempelmeier is with L3S Research Center, Leibniz University Hannover, Appelstraße 9a, 30167 Hannover, Germany 
        {\tt\small tempelmeier@L3S.de}}%
\thanks{$^{3}$Elena Demidova is with Data Science \& Intelligent Systems (DSIS) Research Group, University of Bonn, Friedrich-Hirzebruch-Allee 5, 53115 Bonn, Germany 
        {\tt\small elena.demidova@cs.uni-bonn.de}}%
}

\maketitle

\copyrightnotice

\begin{abstract}
With an increasing number of electric vehicles, the accurate forecasting of charging station occupation is crucial to enable reliable vehicle charging.  
This paper introduces a novel Deep Fusion of Dynamic and Static Information model (DFDS) to effectively forecast the charging station occupation.
We exploit static information, such as the mean occupation concerning the time of day, to learn the specific charging station patterns.
We supplement such static data with dynamic information reflecting the preceding charging station occupation and temporal information such as daytime and weekday.
Our model efficiently fuses dynamic and static information to facilitate accurate forecasting.
We evaluate the proposed model on a real-world dataset containing 593 charging stations in Germany, covering August 2020 to December 2020. 
Our experiments demonstrate that DFDS outperforms the baselines by 3.45 percent points in F1-score on average.
\end{abstract}

\section{Introduction}
\label{sec:intro}

The availability of charging infrastructure for electric vehicles (EV) is vital for adopting EVs within society.  
Recently, the number of EVs has been rapidly growing. 
For instance, in Germany, the number of newly registered EVs in 2020 is three times higher than in 2019 \cite{zdf_ev_numbers}.
The number of EVs and the corresponding charging demand are expected to grow rapidly in the following years.
The number of publicly available charging stations is currently limited, and these stations are not fully used. 
For instance, as of January 2021, only 17,000 public charging stations 
were available throughout Germany \cite{bundesnetzagentur}.
Based on our analysis, these stations are used only 8.8\% of the time, on average.
The data imbalance makes accurate forecasting of charging station occupation difficult.

Accurate forecasting of charging behavior is a relevant problem in many aspects.
First, EV owners can use charging pattern forecasting to identify free charging stations along the route or identify the best time to charge their vehicles.
Second, power grid operators can rely on forecasting to account for the time-dependent energy demand and ensure sufficient power grid capacity.

The forecasting of EV charging behavior is particularly challenging due to several factors.
First, temporal trends such as daytime and day of week substantially influence the charging behavior. 
Moreover, as an EV does not necessarily need to be charged after every trip, the temporal pattern variance is higher than other traffic patterns like, for example, car park occupation.
Second, intuitively, geographic position impacts the charging station occupation. 
Nevertheless, geographic factors that influence charging patterns are widely unexplored.
Third, as of January 2021, due to the currently limited adoption of EVs, 
the existing charging stations are rarely used, as observed in our dataset collected in Germany.
This property leads to skewed data distribution, making accurate forecasting of charging station occupation a challenging task.  
Our experiments demonstrate that existing approaches suffer from data skewness and, as a result, underestimate the charging station occupancy.

In this paper, we propose a novel deep learning approach to forecasting the occupation of specific charging stations.
To this end, we propose an encoder-decoder neural architecture to capture temporal charging patterns.
Furthermore, we employ static station-specific features that capture the typical patterns of individual charging stations and dynamic time features reflecting the current occupation.

In summary, our contributions are as follows:

\begin{itemize}
    \item We introduce the task of EV charging station occupancy forecasting and provide a formalization.
    \item We propose the novel Deep Fusion of Dynamic and Static Information (DFDS) model. DFDS constitutes a deep learning-based approach to forecast individual charging station's occupancy effectively. We propose a novel architecture that effectively combines dynamic and static information.
    \item We conduct an extensive evaluation based on real-world data. Our evaluation shows that we outperform existing approaches by 3.45 percent points in F1-score on average.
\end{itemize}

The rest of the paper is organized as follows: 
Section \ref{sec:problem} presents the problem definition. 
Then in Section \ref{sec:approach} we present the proposed DFDS approach.
Following that, in Section \ref{sec:setup} we describe the evaluation setup.
Then in Section \ref{sec:evaluation} we present the evaluation results. 
We discuss related work in Section \ref{sec:relatedWork}. 
We provide a conclusion in Section \ref{sec:conclusion}.

\section{Problem Definition}
\label{sec:problem}

\begin{figure*}
    \centering
        \includegraphics[width=.75\textwidth]{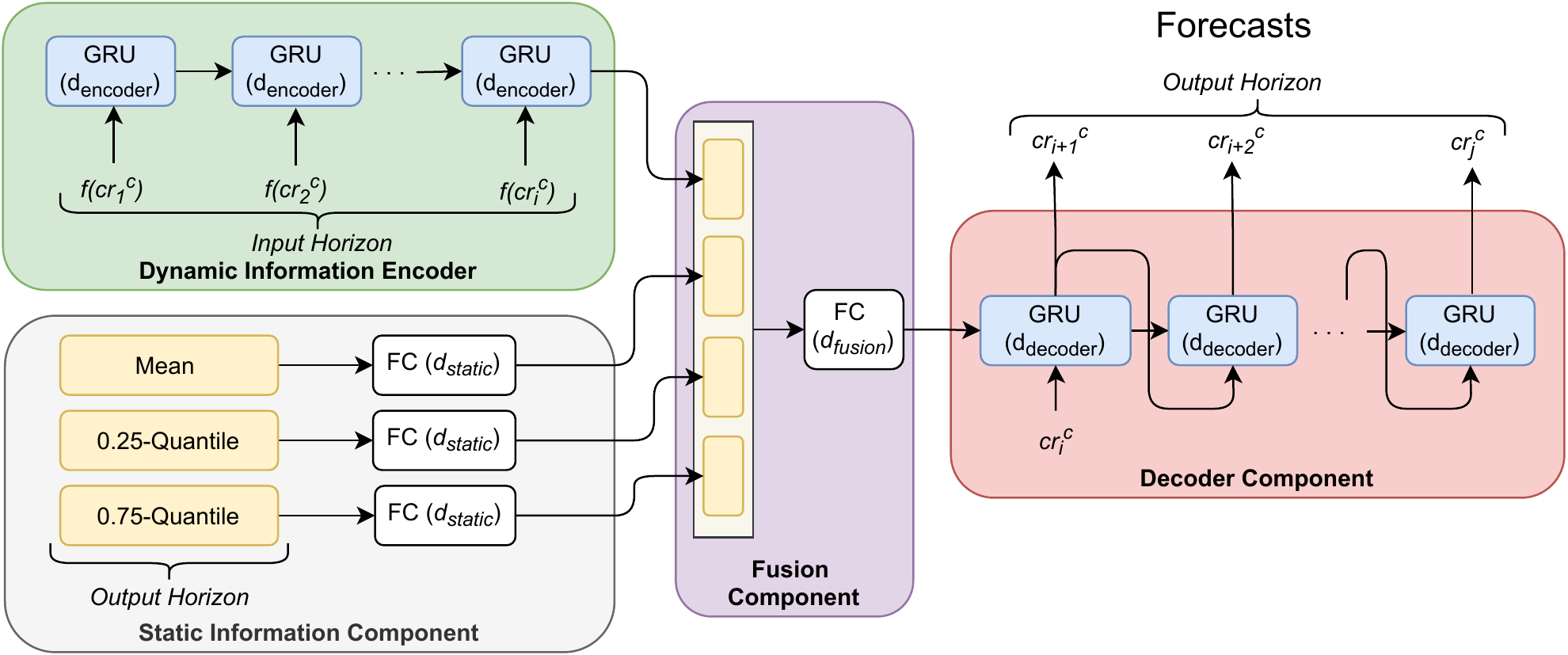}
    \caption{DFDS: Deep  Fusion  of  Dynamic  and  Static information model architecture. 
    The Dynamic Information Encoder represents the GRU-encoder that captures dynamic information.
    The Static Information Component captures the static properties of individual stations.
    The Fusion Component combines the dynamic and static information.
    Finally, the Decoder Component marks the GRU-decoder that makes the forecasts.}
    \label{fig:architecture}
\end{figure*}

In this work, we target the problem of forecasting the occupation of charging stations for electric vehicles. 
In the following, we provide a formal definition of the problem.

Let $C$ denote the set of all charging stations and $T$ the set of all time points.
\begin{definition}[Charging Record]
A \emph{Charging Record} $ cr_t^c \in \{0, 1\}$ provides the occupancy of a charging station $c \in C$ at a time point $t \in T$.
Here, $0$ indicates no occupancy, and $1$ indicates an occupied charging station.
\end{definition}

Temporally consecutive charging records of the same station form a \emph{charging sequence}. A charging sequence is a time series that represents the occupation of an individual charging station over time.

\begin{definition}[Charging Sequence]
A \emph{charging sequence} $s \in S$ for a charging station $c$ consists of charging records $s = \langle cr_n^c, cr_{n+1}^c, ... cr_m^c \rangle$, where $n, n+1, ..., m$ are temporally consecutive time points and $S$ denotes the set of all charging sequences.

\end{definition}

Finally, we define the problem of charging station occupation forecasting as follows:

\begin{definition}[Charging Station Occupation Forecasting]
Given a charging sequence $s$ of length $i$,  ${s = \langle cr_1^c, cr_{2}^c, ..., cr_i^c \rangle}$, for a charging station $c$,
learn a function 
$\hat{y}: S \mapsto S$ that forecasts the occupancy of the same charging station $c$ 
for the next $o$ future time points $\langle cr_{i+1}^c, cr_{i+2}^c, ..., cr_j^c \rangle$, where $o = j -i$.
We call $i$ the \emph{input horizon} and $o$ the \emph{output horizon}.
\end{definition}

\section{Approach}
\label{sec:approach}
An individual charging station's usage is subject to various factors, including its typical usage patterns and the dynamic charging demand. 
Based on this intuition, we propose the  Deep Fusion of Dynamic and Static information (\approach) model, a deep learning approach for the EV charging station occupation forecasting problem.
\approach uses charging record data as training data to learn charging station occupation patterns.
Fig. \ref{fig:architecture} presents the overall architecture of the \approach model.
\approach uses a Gated Recurrent Unit (GRU) \cite{cho-etal-2014-learning} based dynamic information encoder to capture the dynamic occupancy of the charging stations.
Further, we use statistical features to capture the individual station's typical occupation pattern in the static information component.
Finally, we fuse both the dynamic and static information and use a GRU-based decoder to 
forecast charging station usage.
In the following, we detail the individual components of \approach.

\label{sec:architecture}

\subsection{Dynamic Information Encoder}

The dynamic information encoder aims at capturing dynamic patterns of charging station occupancy.
To this end, this component encodes the time-varying attributes of the preceding $i$ time steps denoted as $\langle cr_1^c, cr_2^c, ..., cr_i^c \rangle$.
We employ Gated Recurrent Units (GRU)  to encode the temporal dependencies.
The GRU encoder reads the input $\langle cr_1^c, cr_2^c, ..., cr_i^c \rangle$ in the sequential order, where a charging record $cr_t^c$ at time $t$  is represented by the following features $f(cr_t^c)$:

\textbf{Occupation.} A binary flag that indicates a charging record of the charging station at the time $t$.

\textbf{Day of the week.} Traffic patterns typically differ with respect to the weekday. 
We represent the weekday via one-hot-encoding, i.e. each day of the week is represented as an individual dimension. 
For a particular day, the corresponding dimension is set to '1', while all other dimensions are '0'.

\textbf{Time of day.} The time of day typically influences traffic patterns, for instance, in the form of congestion during rush hours. 
We capture the time of day by one-hot-encoding the hour of the day.
In addition, we round the exact time point to 15-minute intervals and encode the quarter-hours by applying one-hot-encoding.
In total, the time of day is represented by 28 (24 + 4) dimensions.
The separate encoding of the hour of the day and the quarter hours enables the model to ignore the quarter-hour when desirable, e.g., during non-busy hours.

The hidden layer size of the encoder is represented as $d_{encoder}$.

\subsection{Static Information Component}

The static information component aims at capturing the typical patterns of the specific charging stations.
We assume that different charging stations exhibit different typical patterns. 
These usage patterns depend on several factors, e.g., geographic location, charger type or provider. 
For instance, a charging station near a shopping mall might be free in the morning but occupied in the afternoon, while a charging station near an office building might exhibit the reverse pattern.
We capture each station's unique characteristic by aggregating the occupancy at each timestamp of the day from our training set. 
For each time step of the output horizon and each charging station, we compute the following features.

\textbf{Mean Occupation.} The mean occupation of the charging station with respect to the time of the day. 
The feature captures the typical occupation of the charging station.

\textbf{Quantiles.} We compute the 0.25 and 0.75 quantiles of the charging station occupation with respect to the time of the day. 
The quantiles reflect the charging station occupancy patterns during periods of respectively low or high charging demand.

Each feature is fed to a specific fully connected layer with the commonly used ReLU activation function \cite{Goodfellow-et-al-2016}. We denote the size of the fully connected layers by $d_{static}$.

\subsection{Fusion Component}

This component integrates the information from the dynamic information encoder and the static feature component.
To this end, we concatenate the encoded feature vectors and feed them into a fully connected layer with ReLU activation. We denote the size of the fusion layer by $d_{fusion}$.

\subsection{Decoder Component and Prediction}

Finally, we use a GRU-based decoder to predict the occupation of a charging station. 
The decoder component takes the fused feature representation as input. The decoder sequentially predicts the occupation for each time point of the prediction horizon using a fully connected layer with a sigmoid activation function. 
The output of the GRU for each time point acts as an input for the next time point.
This decoding process is recursively applied until the complete sequence for the prediction horizon is generated. 
We denote the hidden layer size of the GRU units as $d_{decoder}$.

We train the \approach model using Binary Cross-Entropy loss as the objective function and the Adam optimizer.

\section{Evaluation Setup}
\label{sec:setup}

In this section, we describe the datasets, evaluation metrics and parameter configurations used in the evaluation.

\subsection{Datasets}
\label{sec:datasets}

We conduct the evaluation on a dataset of EV charging stations located in Lower Saxony, Germany.
The dataset consists of 593 charging stations with 1200 charging outlets and covers August 2020 to December 2020. 
The data was collected during the COVID-19 pandemic. Therefore, we can not determine COVID-19 induced charging behaviour deviations in the dataset.
We collected data of the charging outlet occupancy from a publicly available web service every 15 minutes.
Since we obtain the occupancy information at the outlet level, we evaluate the occupancy prediction for each individual outlet.
This dataset provides occupancy information only.
We consider the data from August 1st to November 15th as the training set (15 weeks) in our experiments.
We create five distinct test sets from the following five consecutive weeks, where each week represents its own test set.

\subsection{Baselines}
\label{sec:baselines}
We compare our approach against the following state-of-the-art baselines. 
For the machine learning and deep learning baselines, we use the occupancy of the past $i$ timestamps as features.
Please note that regression models such as ARIMA are not applicable, as this paper addresses a classification problem.\\
\textbf{Heuristic Baselines.}\\
\textsc{Historical Average}: A na\"ive historical average baseline that strictly predicts the average occupancy with respect to weekday and day time of each charging outlet.\\
\textbf{Machine Learning Baselines.}\\
     \textsc{k-Nearest Neighbors}: A k-Nearest Neighbor classifier. k-NN is a commonly used model for EV charging station demand forecasting. Following \cite{6938864, 6966759, 7297504}, we set $k=1$.\\
     \textsc{Logistic Regression}: A Logistic Regression model that is commonly used for linear classification problems.\\
     \textsc{Random Forest}: A Random Forest model that utilizes an ensemble of uncorrelated decision trees to facilitate non-linear predictions. Random Forests have been used by \cite{7297504} to predict EV charging demand.\\
      \textsc{Support Vector Machine}: A linear support vector machine used by \cite{MAJIDPOUR2016134} to predict EV charging behaviour. Due to numerous training examples and the high computational costs, we apply the bagging \cite{10.1023/A:1018054314350} strategy.\\
\textbf{Deep Learning Baselines.}\\
     \textsc{GRU+Fully Connected}: This model proposed by \cite{Zhu_2019} consists of a Gated Recurrent Unit (GRU)-based encoder to capture the temporal dependencies, followed by a fully connected (FC) layer to predict the occupancy at the next timestamps.\\
     \textsc{Sequence2Sequence}: We extend \cite{Zhu_2019} to a sequence-to-sequence architecture by replacing the fully connected layer with a GRU-based decoder layer. 
    Sequence-to-sequence models constitute the state-of-the-art neural architecture for time series forecasting problems\cite{Goodfellow-et-al-2016}.

\subsection{Evaluation Metrics}
\label{sec:metrics}
To evaluate the performance of the different models, we compute the
following metrics:

\textbf{Precision}: The fraction of the correctly classified instances of occupied charging outlets among all instances that were classified as occupied.

\textbf{Recall}: The fraction of the correctly classified instances of occupied charging outlets among all occupied charging outlet instances.
	
\textbf{F1-score}: The harmonic mean of recall and precision.
We consider the F1-score to be the most relevant metric since it reflects 
both recall and precision.
We report the macro averages of all test sets.

\subsection{Parameter Configuration}

In our experiments, we set the input horizon $i$ to 16 hours, the output horizon $o$ to 8 hours and compute the static features on the entire training set.
We train all deep learning models with the Adam optimizer at an initial learning rate of 0.001 and 20 epochs, after which all models converged.
In our experiments, we observed the best performance of all deep learning models with a hidden layer size of 100.
Thus, we set the hidden layer sizes of the baseline models as well as \approach ($d_{fusion}, d_{static}, d_{encoder}$, and $d_{decoder}$) to 100. 

\section{Evaluation}
\label{sec:evaluation}

The evaluation aims to assess the forecasting performance of the proposed \approach model. 
We further aim to investigate the importance of the features described in Section \ref{sec:approach}.
Finally, we analyze the impact of the input horizon length.

\subsection{Forecasting Performance}
Table \ref{tab:performance} presents the overall forecasting performance of \approach as well as the baselines described in Section \ref{sec:baselines}.
We observe that \approach achieves the best performance in terms of recall (64.53\%) and F1-score (68.55\%) and the third-best performance with respect to precision (73.12\%).
The \textsc{Logistic Regression} baseline achieves the best precision (79.9\%).
However, this baseline suffers from low recall (51.26\%), ultimately resulting in a low F1-score (62.45\%).

Regarding the baselines, the \textsc{Historical Average} baseline achieves the lowest F1-score (26.37\%). 
The low F1-score indicates that simple heuristics are not capable of capturing complex charging patterns.
The machine learning baselines, i.e, \textsc{k-Nearest Neighbours}, \textsc{Random Forest}, and \textsc{Logistic Regression} can improve on the precision but suffer from low recall (50.2\% - 51.2\%).
While \textsc{Random Forest} and \textsc{Logistic Regression} achieve high precision of over 79\%, low recall indicates that these models predict only positive labels with high confidence.

The deep learning models, i.e., \textsc{GRU+Fully Connected} and \textsc{Sequence2Sequence} achieve the highest F1-scores among the baselines (65.05\% and 65.1\%).

Compared to the 
 (\textsc{Sequence2Sequence}) baseline performing best regarding the F1-score, \approach improves on precision (2.01 percent points), recall (4.48 percent points) and F1-score (3.45 percent points).
The \approach model extends the encoder-decoder architecture by using additional dynamic features in the encoder. Further, \approach incorporates additional static features and fuses them with the dynamic information to facilitate the prediction of charging occupation.
We conclude that the dynamic and static information fusion in \approach effectively captures the charging patterns.

\begin{table}
    \centering
    \caption{Overall classification performance for the positive class with respect to Precision, Recall and F1 score [\%]. Best results are marked bold.}
    \label{tab:performance}
    \begin{tabular}{lrrr}
         \toprule
         Approach & Precision & Recall & F1-score  \\
         \midrule
        \textsc{Historical Average} & 46.13 & 18.48 & 26.37 \\
        \textsc{k-Nearest Neighbours} & 64.31 & 50.47 & 56.55 \\
        \textsc{Random Forest} & 79.42 & 50.25 & 61.55 \\
        \textsc{Logistic Regression} & \textbf{79.90} & 51.26 & 62.45\\
        \textsc{Support Vector Machine} & 68.70 & 60.60 & 64.40\\
        \textsc{GRU+Fully Connected} & 71.99 & 59.35 & 65.05 \\
        \textsc{Sequence2Sequence} & 71.11 & 60.05 & 65.10 \\
        \midrule
        \textsc{\approach} & 73.12 & \textbf{64.53} & \textbf{68.55} \\
         \bottomrule
    \end{tabular}
\end{table}

\subsection{Feature Importance}
\begin{table}
    \centering
    \caption{Decrease in prediction performance when a feature is removed [per cent points]. }
    \label{tab:feature_importance}
    \begin{tabular}{lrrr}
         \toprule
         Removed Feature & \makecell{Precision\\ Decrease} & \makecell{Recall\\ Decrease} & \makecell{F1-score\\ Decrease}  \\
         \midrule
        Dynamic Information & 5.27 & 2.32 & 3.67 \\
         \midrule
         Occupation & 38.62 & 20.14 & 29.44\\
         Day of the week & 0.35 & 0.82 & 0.63 \\
         Time of day & 0.65 & 0.37 & 0.53 \\
         \midrule
         Static Information & 1.81 & 4.57 & 3.42 \\
         \midrule
         Mean & 2.32 & 0.27 & 0.91 \\
         0.25-Quantile & 0.32 & 0.01 & 0.14 \\
         0.75-Quantile & -0.13 & 1.61 & 0.86 \\
         \bottomrule
    \end{tabular}
\end{table}

We investigate the individual contribution of each feature to the EV charging station occupancy forecasting problem.
To this end, we carried out a series of experiments where we remove individual features from the \approach model and measure the difference in the forecasting performance.
Table \ref{tab:feature_importance} presents the differences with respect to precision, recall and F1-score in percent points.
The rows \emph{Dynamic Information} and \emph{Static information} correspond to removing the entire respective components from the model.
We observe a decrease in F1-score for all components and features. Hence, all features and components provide valid contributions to the model.

Regarding the dynamic information component, we observe a particular high contribution to the precision (5.27 percent points). 
The high difference in precision indicates that the dynamic information can help to prevent the model from predicting false positives, i.e., predicting an occupied charging station when the station is not used.
Among the dynamic information, the occupation feature has the highest contribution in the F1-score (29.44 percent points). 
Interestingly, removing the occupation feature alone results in a higher difference than removing the entire dynamic information component.
This result indicates that the occupation constitutes the vital information of this component. %
Without the occupation, the additional dimensions introduced by the dynamic information component hinder the model more than helping it.
However, with the occupation feature present, we observe positive contributions in the F1-score also from the other features, i.e. day of the week (0.63 percent points) and time of day (0.53 percent points).
We conclude that the combination of occupation, day of the week, and time of day poses an effective way to incorporate dynamic information.

Moving on to the static information component, we observe an especially high contribution to the recall (4.57 percent points).
This result indicates that the static component helps to mitigate false negatives, i.e., predicting that a station is free when this station is actually occupied.
Among the individual static features, the 0.75-quantile provides the highest contribution to the recall (1.61 percent points) but hurts the precision (0.13 percent points decrease).
Intuitively, the 0.75-quantile overestimates the charging station usage and results in more positive class predictions. 
However, the increased recall results in a positive overall contribution in the F1-score (0.86 percent points).
Furthermore, the mean and 0.25-quantile features increase precision (2.32 and 0.32 percent points). 
These features compensate for the precision decrease introduced by the 0.75-quantile and lead to an overall improvement of precision for the static information component (1.81 percent points).

In summary, we observe that the combination of dynamic and static information enables the effective forecasting of charging station occupancy. The dynamic information is particularly beneficial for precision, while the static information improves recall.

\subsection{Input Horizon Length Impact}
\begin{figure}
    \centering
        \includegraphics[width=.45\textwidth]{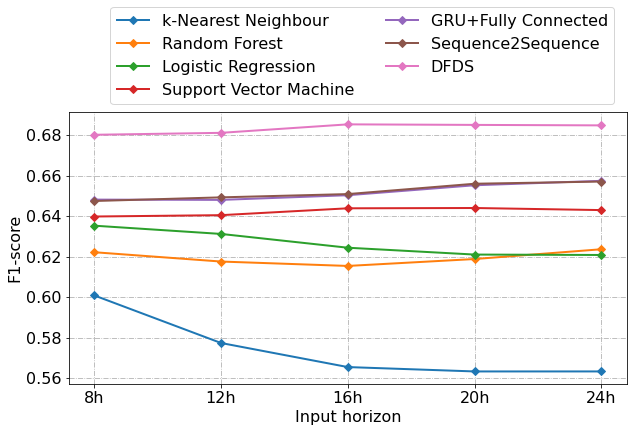}
    \caption{Effect of input horizon length on \approach and baseline models with respect to F1-score.}
    \label{fig:inputHorizon}
\end{figure}

We investigate the impact of the input horizon length by varying the horizon from 8 hours to 24 hours.
Fig. \ref{fig:inputHorizon} presents the F1-score with respect to the input horizon  of  our proposed \approach model and the baselines (except  the input horizon-independent \textsc{Historical Average} model).

The \textsc{k-Nearest Neighbors} shows a decrease of F1-score with an increasing input horizon, from 60\% (8h) to 56\% (24h). 
We observe similar but weaker trends for the \textsc{Logistic Regression} baseline, for which the F1-score decreases from 63\% (8h) to 62\% (24h). 
The decrease indicates that recent information is most important for these baselines. With an increasing horizon, especially the \textsc{k-Nearest Neighbors} baseline cannot distinguish between recent and older input data.

The \textsc{Support Vector Machine} and \textsc{Random Forest} baseline show a near constant performance around 64\% and 62\%, respectively.
While increasing the input horizon does not degrade the prediction performance of these traditional machine learning models, the baselines cannot exploit the additional information provided by the longer input horizon.

The deep learning baselines \textsc{GRU+Fully Connected} and \textsc{Sequence2Sequence} show a slight improvement in F1-score with an increasing input horizon from 65\% to 66\%.
This result illustrates the capability of deep learning models to exploit large amounts of data effectively.
We observe a similar trend for our \approach model, which also is a deep learning model.
Whereas the current differences are relatively small (around one percent point F1-score), we believe that these observations indicate a promising direction to investigate deeper model architectures (i.e., networks with a higher number of hidden layers) capable of leveraging more data in the future work.

\section{Related Work}
\label{sec:relatedWork}
This section discusses related work in charging station demand forecasting and spatio-temporal prediction.

\textbf{EV Charging Station Demand Forecasting}:
Currently, only a few studies address the problem of EV charging station demand forecasting.
Majidpour et al. early investigated the problem in small geographic areas, i.e., a university campus \cite{7297504,MAJIDPOUR2016134}.
They found the use of traditional machine learning models, such as Random Forest, k-Nearest Neighbours, and Support Vector Regression models, to be the most effective to facilitate predictions. 
We modified these models as per the classification problem and compared our approach with those as baselines.
%We compare our approach to these models as baselines.

Another line of research focuses on predicting aggregated power consumption of EV charging stations at the power grid level.
The existing literature discusses a variety of models including analytical model \cite{8917359}, linear regression \cite{li2013}, ARIMA models \cite{doi:10.1080/15325008.2017.1336583, 8764110}, machine learning models \cite{en13164231,8764110}, as well as deep learning models \cite{en12142692,en14051487}.
While these approaches focus on the overall power consumption by EV charging stations, this paper addresses the problem of occupation forecasting for individual charging stations.

Another popular area of research focuses on the planning of charging infrastructure \cite{8917284, 6958139}. Although this is an active area of research, the existing studies do not consider charging demand forecast techniques, which could be a crucial factor for effective charging infrastructure planning. 

\textbf{Spatio-Temporal Prediction}
Several studies investigated various aspects of spatio-temporal prediction. 
This section discusses approaches most relevant to transportation. Wang et al. provide a general discussion in a recent survey \cite{9204396}.

Spatio-temporal demand predictions have been studied in a variety of domains, including
crowd flow prediction \cite{10.5555/3298239.3298479}, bike flow prediction \cite{10.1145/2971648.2971652,  10.1016/j.pmcj.2010.07.002}, and taxi demand prediction \cite{7795558, 8569427}.
Existing approaches either use \textit{cluster-based}, \textit{grid-based} and \textit{instance-level} models.

Cluster-based approaches first determine clusters with respect to distance and usage information. Then, the models jointly predict the demand of each cluster \cite{10.1145/2971648.2971652,  7795558}.
These approaches typically suffer from imprecision for specific instances induced by the clustering.

Grid-based approaches divide a geographic area into fixed-sized cells (areas). 
Such discretization enables the use of popular deep learning models such as convolutional neural networks \cite{10.5555/3298239.3298479, 8569427}.
However, the arbitrary discretization of the grid hinders these models in fine-granular predictions.

Instance-level approaches enable the prediction for the individual instances, e.g., specific bike-sharing stations.
Compared to cluster-based and grid-based approaches, instance-based approaches are more complex as they require predictions for each instance. 
Typical instance-based models include time series analysis \cite{10.5555/2045723.2045734, 10.1109/MDM.2012.16} and statistical models \cite{10.1016/j.pmcj.2010.07.002}.
In this work, we proposed a novel deep learning method to forecast the instance-level charging station occupation.

\section{Conclusion \& Future Work}
\label{sec:conclusion}

In this paper, we addressed the problem of forecasting electric vehicle charging station occupation.
We proposed the \emph{Deep Fusion of Dynamic and Static information} (DFDS) model, a deep learning approach that effectively combines dynamic and static information.
\approach exploits the typical static patterns of the individual charging stations, such as regular occupation rates, and the dynamic information, such as the current occupation, to facilitate occupation predictions.
Our experimental evaluation shows that \approach is highly effective and improves over existing baselines by up to 3.45 percent points in F1-score on average.
Further, we observe that dynamic information helps achieve high precision, while static information enables a high recall.

While \approach is currently limited to predict the occupancy for each charging outlet individually, we would like to investigate the inter-dependencies of outlets at the same charging station in future work.
Furthermore, we would like to explore the effect of additional features such as charging price or waiting time on \approach.
Finally, it would be interesting to investigate the transferability of \approach for other geographic areas and within particular contexts, such as large-scale public events or holiday seasons.

\section*{Acknowledgment}
This work is partially funded by the BMWi, Germany under ``d-E-mand'' (grant ID 01ME19009B) and ``CampaNeo'' (grant ID 01MD19007B), the European Commission (EU H2020, ``smashHit", grant-ID 871477), and DFG, German Research Foundation (``WorldKG'', DE 2299/2-1). 

%\balance

\bibliographystyle{IEEEtran}
\bibliography{refs}

\end{document}